\documentclass{article}

\usepackage{times}
\usepackage{graphicx} 
\usepackage{subfigure} 

\usepackage{natbib}

\usepackage{algorithm}
\usepackage{algorithmic}
\usepackage{mathtools}
\usepackage{amsmath,amssymb}
\usepackage{authblk}

\usepackage{hyperref}

\begin{document} 
\title{Information gain ratio correction: Improving prediction with more balanced decision tree splits}
\author[1]{Antonin Leroux}
\author[1]{Matthieu Boussard}
\author[1]{Remi D\`es}
\affil[1]{craft ai}
\maketitle





\begin{abstract} 
Decision trees algorithms use a gain function to select the best split during the tree's induction. This function is crucial to obtain trees with high predictive accuracy. Some gain functions can suffer from a bias when it compares splits of different arities. Quinlan proposed a gain ratio in C4.5's information gain function to fix this bias. In this paper, we present an updated version of the gain ratio that performs better as it tries to fix the gain ratio's bias for unbalanced trees and some splits with low predictive interest.
\end{abstract}

\section{Introduction}
\label{ap: intro}
Decision Trees (DT) classify data by dividing the feature space into subregions whose bounds are given by the DT. The tree structure does this by recursively dividing into different branches the samples present in each node. All leaves are labeled and the samples in the final subregion corresponding to the leaves are classified accordingly.

The most popular DT classification algorithms like CART \cite{Breiman}, C4.5 \cite {Quinlan} and OC1 \cite{murthy1993} generate a top-down DT by applying recursively a splitting method on the data. At each node the splitting algorithm is applied on the remaining data to find the optimal decision rule. The algorithm stops when the sub-space reached is pure or almost pure. In all of those algorithms, the slitting methods plays a critical role.

There are a lot of different splitting methods. However, their aim is always the same: dividing an original dataset into smaller groups, where the number of groups depends on the chosen splitting algorithm. In order to perform such a division, the algorithms need a gain function to assess the quality of the different splits. 

The main contribution of this paper is to propose a new gain function, derived from the one used by \citeauthor{Quinlan} in C4.5, that performs better as it tries to fix the gain ratio's bias for unbalanced trees and some splits with low predictive interest.

The remaining sections of this paper are organized as follows: Section \ref{sec:relatedWork} presents related work. Section \ref{ap:gain} explains the gain functions and the contribution of this paper, a modified gain function, called the balanced gain ratio. Section \ref{seq:result} presents a comparison between the balanced gain ratio and the C4.5 gain ratio on several datasets. Section \ref{seq:conculsion} concludes this article.

\section{Related Work}
\label{sec:relatedWork}
In this section we give a quick review of existing gain functions. \cite{Breiman} and \cite{Quinlan} introduced gain functions based on the measure of the impurity of the nodes. The important criterion is then the purity of the subregions created. Therefore a good split will be one that creates partitions composed of mainly one class. Impurity-based criteria thus became the norm for gain functions. The Gini index is the criterion used in \citeauthor{Breiman}'s CART and the Gain ratio proposed in \cite{Quinlan} is derived from the information gain. Information gain and Gini are very similar functions. We will give more details about them in Section \ref{ap:gain}. 

Other impurity-based criteria exist, trying to correct a specific bias or target specific problems. The Likelihood Ratio Chi-squared statistics introduced in \cite{Attneave1959} is measuring the statistical significance of the information gain criterion. The aim is to evaluate the dependence of the target feature and the input . The DKM criterion \cite{kearns89} has been proved to require smaller trees than Gini or the information gain to reach a certain error.

Unlike the Gini index or the two criteria mentioned above, some criteria are "normalized" meaning that a correction is brought to an original impurity measure (like \citeauthor{Quinlan}'s gain ratio), the reasons for this normalization are given below in Section \ref{ap:gain}. That is the case of the Distance Measure \cite{Lopezdemantra1991}, it normalizes the goodness-of-split measure \cite{gain_review} in a similar way that the gain ratio does for the information gain. There is also the Orthogonal criterion from \citeauthor{fayyadirani1992}, it works on the angle of vectors, that represent the probability distribution of the desired attributes in the partitions given by the split. It is a criterion for binary split but has been shown to perform better than both Gini and information gain for specific problems. Another binary gain function is the Kolmogorov-Smirnov distance criterion proposed in \cite{Friedman1977}. This test evaluate a difference between probabilities of good classification for each class of the current split. Area Under the Curve (AUC) metric is used to evaluate the performance of classifiers, but according to \cite{Ferri2002}, selecting the split that has maximal area under the convex hull of the ROC curve can be used as a splitting criterion. This criterion has shown good results in comparison to other methods.

\section{Gain Function}
\label{ap:gain}
Here we give further explanations on the choice of the gain function that is proposed in this article.

\subsection{Impurity measures and Gain functions}
\label{sec:impurity}
The impurity measures are used to estimate the purity of the partitions induced by a split. For the total set of samples $C$ we have $K$ classes. We consider a split $S$ that creates $J$ partitions $C^j$ of the samples. We note $n_{k}=|C_k|$ and $n^{j}=|C^j|$, respectively the number of samples of class $k$ and the number of samples in the partition $j$. We define the number of samples of a given class present in a given partition as:
\begin{equation}
n_{k}^{j} = |C_k \cap C^j | ,\quad \forall j \in [1,..,J], k \in [1,..K]
\end{equation}
And finally the proportion of class $k$ in partition $C^j$: 
\begin{equation}
p_{k}^j = \frac{n^{j}_{k}}{n^{j}}
\end{equation}

We can now define the gini index Eq.\ref{eq:gini} and entropy Eq.\ref{eq:entropy} impurity measures of partition $C^j$. 
\begin{equation}
\label{eq:gini}
I(C^j)=G(C^j) = - \sum^{K}_{k=1} (p_k^{j})^2
\end{equation}
\begin{equation}
\label{eq:entropy}
I(C^j)=E(C^j) = - \sum^{K}_{k=1} p_k^{j}\textnormal{log}(p_k^{j})
\end{equation}
We can also define $I(C)$ for the whole data set by replacing $p_k^j$ by $p_k=\frac{n_k}{n}$. Then we have the purity Gain function of splitting $C$ according to $S$: 
\begin{equation}
\label{eq:gain general}
G(S,C)=I(C) - \sum^{J}_{j=1} \frac{n^j}{n}I(C^j)
\end{equation}
\citeauthor{Breiman} used that exact gain for the CART algorithm. But \citeauthor{Quinlan} introduced an adjusted gain for the C4.5 algorithm. Indeed, as described in \cite{quinlan1995} there exists some \emph{"fluke theories that fit the data well but have low predictive accuracy"}. The Gain function as written in Eq.\ref{eq:gain general} has a bias toward categorical features with a lot of different values. In the extreme case where $n^j =1$, $\forall j \in [1,..,J]$, the impurity is 0 and the gain is maximal. Such a bias is reached when the samples have a categorical feature (attribute with unordered value, for instance days of the week or country names) that has a lot of possible values (possibly one for each sample).

To fix this issue \citeauthor{Quinlan} decided to use a split information coefficient. With $p^j=\frac{n^j}{n}$ it is defined as: 
\begin{equation}
\label{eq:split info}
\textnormal{SplitInformation}(S,C)=-\sum^{J}_{j=1} p^{j}\log(p^{j})
\end{equation}
With this we can now define the gain ratio used in C4.5 as:
\begin{equation}
\label{eq:gain ratio}
\textnormal{GainRatio}(S,C) = \frac{G(S,C)}{\textnormal{SplitInformation}(S,C)}
\end{equation}
We can show that the split information helps to solve the bias. It can be rewritten as:
\begin{equation}
\label{split info}
\textnormal{SplitInformation}(S,C) = \log(n) - \sum^{J}_{j=1} p^j \log(n^j)
\end{equation}
And so is maximal when the sum is null (i.e. when $n^j=1, \forall j \in [1,..,J]$). As the total gain decreases when the split information increases, this coefficient brings a good correction for the bias described earlier. 

\subsection{Balanced Gain ratio}
\label{sec:proposed gain}
But this adjustment brings a bias of its own. Indeed, the split information is actually an impurity measure of the partitions. That is why the gain ratio of C4.5 will be biased toward splits that produce a subnode with a huge part of the total samples. In the case of binary splits it will then have a tendency to isolate a quite pure and small part of the samples in one side while keeping the major part of the samples in the other node. 

This is problematic as the algorithm will produce unbalanced trees that require more depth to obtain a good accuracy. Such trees take longer to compute and are less human readable. This behavior can also damage the prediction accuracy as the algorithm focuses more on the samples the furthest from the center. Those samples may be exceptional values or errors that often prove to be less interesting in the generalization process. 

Thus, we propose a method that diminishes this adjustment brought by the split information, while conserving it for the reasons we described earlier in this section. 

The new adjusted gain ratio function is: 
\begin{equation}
\label{gain}
\Gamma(S,C)= \frac{G(S,C)}{1+\textnormal{SplitInformation}(S,C)}
\end{equation}

We want to remove the undesirable behavior of the gain ratio, described in the beginning of \textbf{\ref{sec:proposed gain}}, while keeping the correction introduced by \citeauthor{Quinlan}. The bias toward categorical feature with a lot of values is corrected when the split information reaches high values. The bad behavior appears when the split information takes small values and helps overly unbalanced splits obtain a high gain. That is why we brought the $1 + $ correction. It preserves the global monotony of the adjustment coefficient but attenuates the modification when the split information is small. 

\section{Results}
\label{seq:result}
In this section we present empirical results to evaluate the performance of the corrected gain ratio. We test our algorithm with real data sets coming from the UCI repository \cite{uci}. All estimations are made with 10 5-fold cross validations to estimate the average accuracy of the produced tree. The produced trees are then pruned using pessimistic error pruning. Some of the chosen datasets have categorical features while some others are fully numerical.

The Table \ref{tab: data continuous} shows the values of $d$ (number of features), $n$ (number of samples) and $K$ (number of classes) for the different datasets used. The $d_c$ value is the number of categorical attributes.

\begin{table}[htbp]
\caption{Real continuous datasets downloaded from UCI.}
\label{tab: data continuous}
\vskip 0.15in
\begin{center}
\begin{small}
\begin{sc}
\begin{tabular}{lcccr}
\hline
Data set & $d$($d_c$) & $K$ & $n$ \\
\hline
Heart    & 13(0) & 2 & 270  \\
\hline
Indian Diabetes (PIMA) & 8(0) & 2 & 768 \\
\hline
Glass    & 9(0) & 6 & 214 \\
\hline
Wine &13(0) &  3 & 178 \\
\hline
Survival & 3(0) & 2 &306 \\
\hline
Letter & 16(0) & 26 & 20000 \\
\hline
Livers (BUPA) & 6(0) & 2 & 345 \\
\hline
Balance Scale & 4(0) & 3& 625 \\
\hline
Income & 14(8) & 2 &32561 \\
\hline
Bank & 16(9) & 2 &45211 \\
\hline
\end{tabular}
\end{sc}
\end{small}
\end{center}
\end{table}

\subsection{The algorithm}
\label{subsec: algo}
Since our proposed gain is an improvement over \citeauthor{Quinlan}'s gain for C4.5, we used C4.5 to evaluate it. C4.5 can treat datasets with both categorical and numerical attributes and produces axis-parallel splits. It splits on the feature that produces the highest gain. For numerical attributes, the split is binary whereas for categorical features it depends on the number of values.

\subsection{C4.5 Gain Ratio vs Balanced Gain ratio}
Table.\ref{tab:vs gain} shows the comparison between the Proposed Gain Function and the Gain ratio introduced by \citeauthor{Quinlan}. We ran tests using the algorithm we presented in Subsection \ref{subsec: algo} with the two differents gain functions. The column \emph{Gain Ratio Accuracy} presents the average accuracy obtained with the gain ratio function. The \emph{Balanced Gain Ratio Accuracy} gives the accuracy for the gain proposed in this paper. The last column, \emph{Diff.} is the average difference in accuracy between the two gains.

For the large datasets in number of features, (\emph{Income}, \emph{Bank} and \emph{Letter}), the results are better for the proposed gain. It is consistent with the analysis we provided in \textbf{\ref{sec:proposed gain}} as balancing the tree helps obtaining better results and shorter trees. The trees with lower depth are also much quicker to compute. The difference between the two gains can be very important. For the latter dataset, the depths of the trees produced are around 20 for our proposed gain while for the C4.5 gain ratio they are around 100.

On smaller datasets the proposed gain performs generally better than the classical gain ratio, except for the \emph{survival} and \emph{Heart} datasets. These two datasets are among the smallest with less than 350 samples in them. However for \emph{BUPA}, \emph{Balance} and \emph{PIMA}, we observe an improvement of several percents in accuracy. This shows that sometimes the correction brought by \citeauthor{Quinlan} to the original information gain leads to bad predictive accuracy. This bad behavior is corrected by the factor we added. 

Thus, it is clear that for small datasets that produce small trees, the Proposed Gain is not really necessary even though results are equivalent. On the other hand, the proposed gain brings real improvement for bigger datasets.

On datasets with categorical features (\emph{Income} and \emph{Bank}), we could have expected a drop in accuracy as our new gain reduces the correction brought by \citeauthor{Quinlan}. This was particularly important for splits on categorical features (see Sec.\ref{sec:impurity}). Yet the new gain achieves better accuracy every time for both datasets.  
\begin{table}[htbp]
\begin{center}
\caption{C4.5 Gain Ratio vs Balanced Gain}
\label{tab:vs gain}
\vskip 0.15in
\begin{small}
\begin{sc}
\begin{tabular}{lcccr}
\hline
Data set & Gain Ratio & Balanced Gain Ratio & Diff.  \\
 & Accuracy & Accuracy & \\
\hline
Glass  &  64.02& \textbf{65.42}  & +1.4 \\
\hline
Bupa & 60.58& \textbf{66.67} & +6.09 \\
\hline
Heart  & \textbf{78.15} &77.78 & -0.37 \\
\hline

Balance  & 73.92 & \textbf{77.60} & +3.68 \\
\hline
Survival  & \textbf{73.86} & 72.87 & -0.99\\
\hline
PIMA  & 72.79 & \textbf{75.26} & +2.47 \\
\hline
Wine  & 94.38 & 94.38 & +0.00 \\
\hline
Bank & 89.87 & \textbf{90.23} & +0.36 \\
\hline
Income & 83.92 & \textbf{84.26} & +0.34 \\
\hline
Letter & 86.85 & \textbf{87.40} & +0.55\\
\hline
\end{tabular}
\end{sc}
\end{small}
\vskip -0.1in
\end{center}
\end{table}

\section{Conclusion}
\label{seq:conculsion}
In this paper we presented a corrected version of the gain ratio used in the C4.5 algorithm. This new gain function helps to balance trees and improves accuracy. The biggest improvement lies in the depth of the tree produced for big datasets, which comes with a significant improvement of the computation time as well. It proves to be working better than the original gain ratio in most cases. The correction we proposed could also be interesting for other kinds of splits we did not mentioned in this paper, such as non-binary splits on numerical features \cite{multiway}.

\nocite{gain_review}

\bibliography{biblio}
\bibliographystyle{icml2017}

\end{document}